\pdfoutput=1

\documentclass[11pt]{article}

\usepackage{acl}

\usepackage{times}
\usepackage{latexsym}

\usepackage{pgfplots}
\pgfplotsset{every axis/.append style={label style={font=\tiny}, tick label style={font=\tiny}}}

\usepackage[T1]{fontenc}

\usepackage[utf8]{inputenc}

\usepackage{microtype}
\usepackage{multirow}
\usepackage{multicol}

\usepgfplotslibrary{groupplots}
\pgfplotsset{compat=1.16}

\usepackage{todonotes} 

\usepackage{bbm}
\usepackage{array}\newcolumntype{P}[1]{>{\centering\arraybackslash}p{#1}}
\usepackage{makecell}
\usepackage{graphicx}
\usepackage{amsmath}
\usepackage{amssymb}
\usepackage{pifont}
\usepackage{booktabs}
\usepackage{enumitem}
\usepackage{soul}
\sethlcolor{Lavender}
\usepackage{moresize}
\usepackage{multicol}
\usepackage{multirow}
\usepackage[export]{adjustbox}
\usepackage{subfigure}

\definecolor{pink}{rgb}{1, 0.75686, 0.796}
\definecolor{orange}{rgb}{1, 0.647, 0}
\definecolor{gold}{rgb}{1, 0.8392, 0.078}
\definecolor{purple}{rgb}{0.6, 0.4, 0.8}

\title{In-Context Analogical Reasoning with Pre-Trained Language Models}

\author{
Xiaoyang Hu\textsuperscript{$1$$2$\thanks{\hspace{5pt}Authors contributed equally to this work.}}
\hspace{20pt}
Shane Storks\textsuperscript{$1$\footnotemark[1]}
\hspace{20pt}
Richard L. Lewis\textsuperscript{$2$}\thanks{\hspace{5pt}Equal advising contribution.}
\hspace{20pt}
Joyce Chai\textsuperscript{$1$}\footnotemark[2] \\
$^1$Computer Science and Engineering Division, University of Michigan \\
$^2$Department of Psychology, University of Michigan \\
\texttt{\{nickhu, sstorks, rickl, chaijy\}@umich.edu}
}

\begin{document}

\maketitle
\setcounter{footnote}{1}

\begin{abstract}
Analogical reasoning is a fundamental capacity of human cognition that allows us to reason abstractly about novel situations by relating them to past experiences. While it is thought to be essential for robust reasoning in AI systems, conventional approaches require significant training and/or hard-coding of domain knowledge to be applied to benchmark tasks. Inspired by cognitive science research that has found connections between human language and analogy-making, we explore the use of intuitive language-based abstractions to support analogy in AI systems. Specifically, we apply large pre-trained language models (PLMs) to visual Raven's Progressive Matrices (RPM), a common relational reasoning test. By simply encoding the perceptual features of the problem into language form, we find that PLMs exhibit a striking capacity for zero-shot relational reasoning, exceeding human performance and nearing supervised vision-based methods. We explore different encodings that vary the level of abstraction over task features, finding that higher-level abstractions further strengthen PLMs' analogical reasoning. Our detailed analysis reveals insights on the role of model complexity, in-context learning, and prior knowledge in solving RPM tasks.



\end{abstract}

\section{Introduction}\label{sec:intro}
Humans are constantly presented with novel problems and circumstances. Rather than understand them in isolation, we try to connect them with past experiences. With any luck, we might find an \textit{analogy}: a mapping between relevant aspects of this new situation and a past situation, which 
helps form abstractions that allow us to reason more effectively
in the future 
\cite{holyoak1984analogical}. Analogy is thought to underpin humans' robust reasoning and problem solving capabilities \cite{hofstadter2013surfaces}, and thus it is believed to be prerequisite in order to enable the same in AI systems. However, conventional approaches struggle with analogy-making, and are trained on thousands of examples to achieve any success on benchmark tasks. 
 This is
unsatisfying, 
as humans are capable of analogy-making without explicit training, and such analogy-making should
enable zero-shot generalization to new situations \cite{mitchell2021abstraction}.

\begin{figure}
    \centering
    \includegraphics[width=0.48\textwidth]{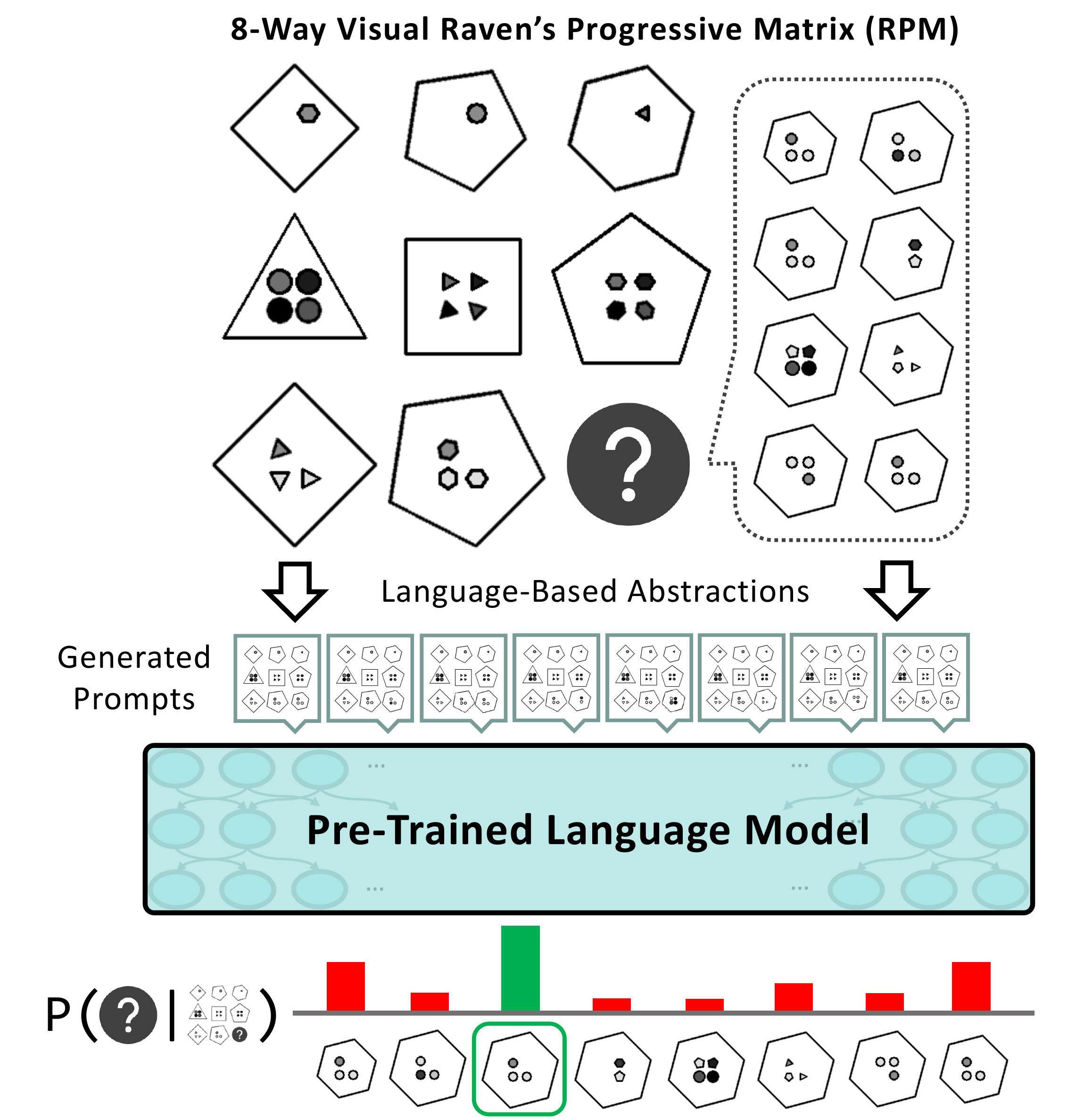}
    \caption{Raven's Progressive Matrices~\cite{raven1938raven,zhang2019raven} are an analogy-making task where one must infer the missing matrix item based on abstract rules instantiated in the first two rows. To demonstrate the potential analogical reasoning skills in pre-trained language models, we develop language-based abstractions over their key perceptual features, then prompt them to select the completion of the matrix.}
    \vspace{-10pt}
    \label{fig: abstractions in rpm}
\end{figure}

Interestingly, a body of work in cognitive science suggests that analogy-making and relational reasoning are connected to humans' symbol system and language capabilities \cite{gentner2010bootstrapping}. For example, \citet{gordon2004numerical} finds that members of an Amazonian tribe that count only with words for ``one,'' ``two,'' and ``many'' struggle to make analogies with higher numbers. Further, \citet{gentner2013spatial} find that deaf children whose sign language does not involve spatial relations are outperformed by hearing children on a spatial relational reasoning task, while \citet{christie2014language} find that assigning even nonsensical names to relations enhances children's relational reasoning. All of this demonstrates that language serves as a powerful way for humans to abstract and better reason about the overwhelming and complex percepts we encounter in the world.

In this work, we explore whether language may serve a similar purpose in AI systems. Specifically, we apply contemporary autoregressive pre-trained language models (PLMs) to Raven's Progressive Matrices (RPM), an example of which is shown in Figure~\ref{fig: abstractions in rpm}. RPM is a widely used psychometric test for relational reasoning that requires inducing an abstract rule from just two examples of short sequences of groups of shapes, and then applying the rule to complete a new partial sequence~\cite{raven1938raven}. This task makes minimal assumptions about the test taker's prior knowledge, and is thus thought to provide a good estimate for
general intelligence \cite{holyoak2012analogy}.
On the RAVEN dataset~\cite{zhang2019raven}, we find that given the ability to perceive key features of RPMs, large PLMs exhibit a surprising capacity for zero-shot relational reasoning, approaching that of supervised vision-based deep learning approaches and even humans. We propose three levels of abstraction over the language features of the task using name assignment and task decomposition, and find that each abstraction further strengthens PLMs' relational reasoning.
Our results and detailed analysis offer insights on PLM performance, including the role of models' complexity, in-context learning, and prior knowledge in emergent relational reasoning, and suggest that they could play an important role in future cognitive architectures for analogy-making.\footnote{Experiment code is available at \url{https://github.com/hxiaoyang/lm-raven}.}

\section{Related Work}
Past work has studied analogy in AI across various domains. \citet{mitchell2021abstraction} provides a comprehensive overview of these efforts, especially those applied in idealized symbolic domains. Here, symbolic and probabilistic methods have traditionally been applied \cite{gentner1983structure,hofstadter1994copycat,lake2015human}. However, these approaches typically require hard-coding domain-specific concepts, and require substantial search through domain knowledge to operate on their target problems, thus making them unscalable. The creation of large-scale image datasets for analogy tasks here \cite{zhang2019raven,hu2021stratified,odouard-2022-arc} have enabled further research with deep learning and neuro-symbolic methods 
\cite{hill2019learning,spratley2020closer,NEURIPS2020_c39e1a03,Zhang_2021_CVPR}, which bring the advantage of requiring less ad-hoc encoding of domain knowledge, but require thousands of training examples to learn the tasks, still limiting their generalization capability.

Other work has explored AI systems' analogy-making in real-world domains, including in natural images~\cite{teney2020v,bitton-2022-vasr} and language~\cite{li-etal-2020-ca,chen-etal-2022-e,sultan-2022-circus}, especially lexical analogies \cite{turney2003combining,turney2008latent,speer2008analogyspace,mikolov-etal-2013-linguistic,mikolov2013distributed,linzen-2016-issues,lu2019emergence}. However, these domains make it difficult to control the prior knowledge required to solve tasks \cite{mitchell2021abstraction}, and in the context of recent generative foundation models that are extensively pre-trained on natural data, it becomes difficult to separate analogy learning from distributional patterns that can be overfit.
Unlike prior work, we apply such foundation models for language to analogical reasoning in a zero-shot setting, bypassing the requirement of hard-coding domain knowledge or training models on task-specific data.
{Furthermore, while contemporaneous work has applied PLMs to a variety of simpler relational reasoning tasks in language \cite{webb2022emergent}, we systematically explore the advantage of using language to abstract over complex visual features of the task, opening questions about how the powerful symbol systems learned in PLMs may support robust, perception-driven reasoning in future AI systems.}

\section{Raven's Progressive Matrices}

Raven's progressive matrices (RPM) are abstract relational reasoning tasks used in cognitive psychology to test humans' analogy-making \cite{raven1938raven}.
Each instance of RPM is a matrix consisting of 9 \textit{items} arranged in a square, the last of which must be selected from a set of choices. Each item consists of several perceptual \textit{attributes}, such as shape, color, or more abstract features. Within each row of the matrix, a \textit{relation} is applied over these attributes, such as progression of numerical values associated with these attributes. Given the first two rows of the matrix, the challenge of the task is to identify the relations being applied to items, and apply them analogously in the third row to infer the missing ninth item. Successfully solving an RPM requires tackling two sub-problems: \textit{perception} of each item's attributes, and \textit{reasoning} over multiple items' attributes to infer and apply relations.

\subsection{RAVEN Dataset}\label{sec:datasets}
\begin{figure}
    \centering
    \includegraphics[width=0.485\textwidth]{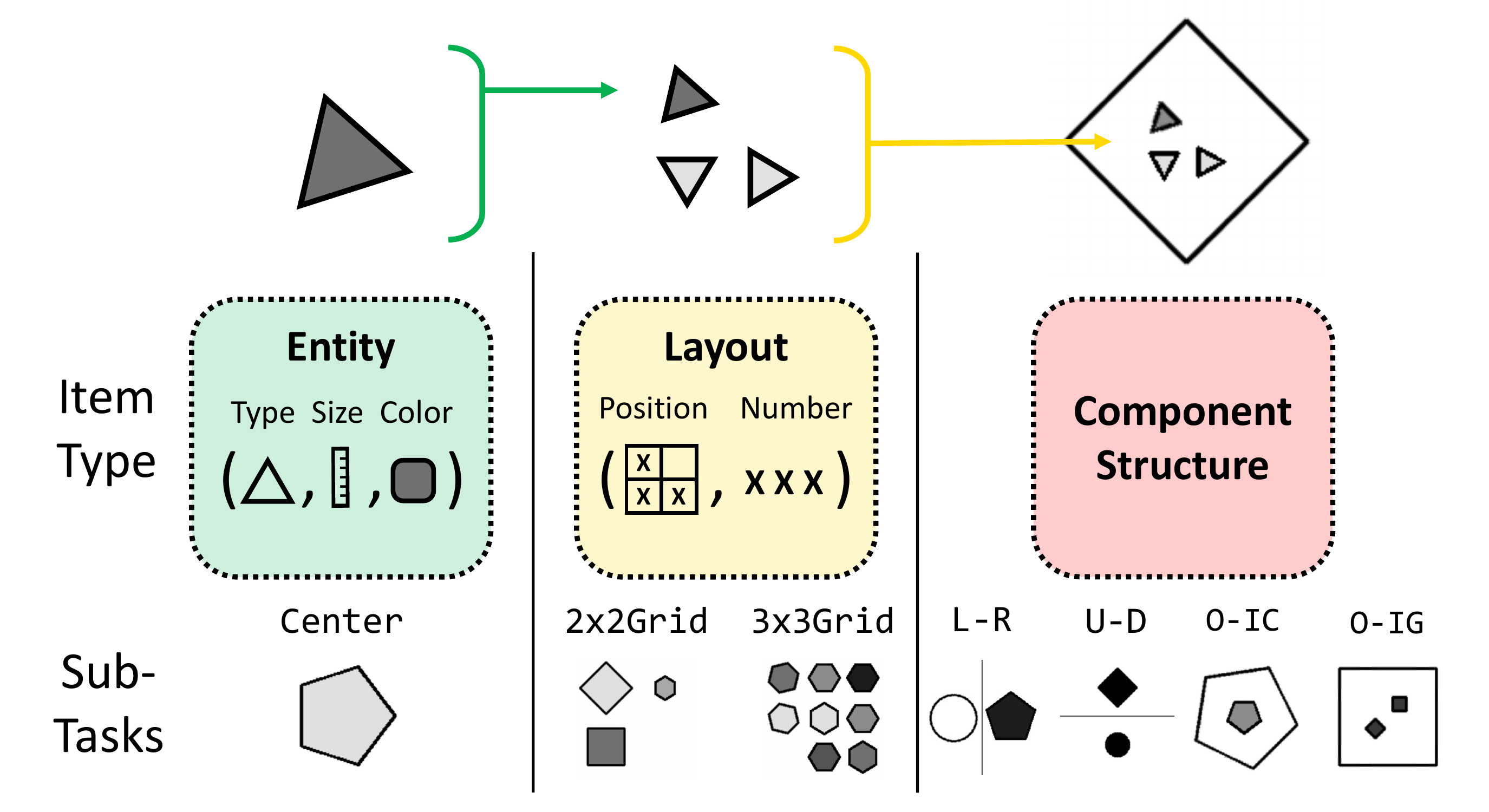}
    \vspace{-15pt}
    \caption{Illustration of the compositional nature of entities, layouts, and component structures in RAVEN, and their unique attributes. We provide example items from sub-tasks each item type appears in.}
    \vspace{-10pt}
    \label{fig: entity layout structure}
\end{figure}

We focus our study on RAVEN~\cite{zhang2019raven}, which provides a large-scale benchmark for RPM tasks for training and evaluation of AI systems. 
Each RPM has 8 possible candidate items to complete it.
As shown in Figure~\ref{fig: entity layout structure}, each item may consist of compositional \textit{entities}, \textit{layouts}, and/or \textit{component structures}, and RAVEN provides a suite of increasingly complex sub-tasks built from these elements. We introduce their unique attributes below, as well as relations that may occur over them across items in the matrix.

\paragraph{Entities.}
A single entity has a \texttt{type} (i.e., shape), \texttt{size}, and \texttt{color} selected from a small number of classes. Each of these attributes is associated with a number: \texttt{type} with the number of sides in the entity's shape, \texttt{size} with its diameter, and \texttt{color} with the darkness of its shading. The simplest sub-task of RAVEN is \texttt{Center}, where each item only consists of a single entity. 

\paragraph{Layouts.}
Layouts of entities bring additional higher-level attributes to items, specifically the \texttt{number} (i.e., count) and \texttt{position} of entities within a layout.
In the \texttt{2x2Grid} and \texttt{3x3Grid} sub-tasks of RAVEN, each item consists of multiple entities arranged in a grid.

\paragraph{Component structures.}
Items may also be composed of multiple sub-items or \textit{components}; RAVEN includes four sub-tasks that introduce this even higher-level challenge: \texttt{L-R}, \texttt{U-D}, and \texttt{O-IC}, each of which consist of two single entities in different configurations, and \texttt{O-IG}, which consists of a 2-by-2 grid inside of a larger entity.

\paragraph{Relations.}
Following prior work on this task, RAVEN applies four different relations to item attributes across rows of the matrix. These are \texttt{Constant}, which does not modify an attribute, \texttt{Progression}, which increases or decreases the value of an attribute by 1 or 2, \texttt{Arithmetic}, which performs addition or subtraction on the first two attributes of the row to create the third, and \texttt{Distribute Three}, which distributes three consistent values of an attribute across each row.

\section{Methods}

In order to apply PLMs to RAVEN, we abstract the visual features of the task into language.
Our abstractions are intentionally applied on a per-item basis to tackle the perception problem of the task without giving the PLM explicit hints toward the reasoning problem (which requires capturing patterns over multiple items). This allows us to focus on evaluating the reasoning capabilities of PLMs.\footnote{As the important features of RAVEN are simple, the perception of an individual item is better performed by computer vision models, and can already be done to fairly high accuracy \cite{Zhang_2021_CVPR}. For more general-purpose analogy-making beyond idealized domains, the robust perception of key features that allow previous (source) experiences to be mapped to novel (target) experiences is a challenging unsolved problem~\cite{mitchell2021abstraction}.}

First, we introduce our multi-level abstractions for the RAVEN dataset.\footnote{Some example PLM prompts using these abstractions are shown in this section, while more examples are provided in Appendix~\ref{apx: example prompts}.} Then we formally define the interface between PLMs and the RPM task.

\subsection{Abstractions in RAVEN}
We define abstractions for entity-level attributes, layout-level attributes, and component structures which convert the RPM task into one or more text prompts. 
We apply two kinds of abstractions: \textbf{naming} and \textbf{decomposition}. As discussed in Section~\ref{sec:intro}, assigning names to perceptual features strengthens humans' analogy-making skills over them. 
Inspired by this, naming abstractions abstract over attributes or combinations of attributes in the RPM by assigning a unique name to describe them. Meanwhile, jointly understanding and tracking the complex features of the task can become a burden even for humans. Inspired by humans' capability to decompose complex tasks into independent sub-tasks~\cite{lee2001does}, decomposition abstractions split the RPM into multiple sub-matrices by its independent features, then generate a separate prompt for each one. We can then prompt a PLM once for each sub-matrix, and 
aggregate PLM outputs to choose a candidate matrix completion.\footnote{A more formal definition for decomposition is provided in Section~\ref{sec:formal definitions}.}

\begin{figure}
    \centering
    \includegraphics[width=0.49\textwidth]{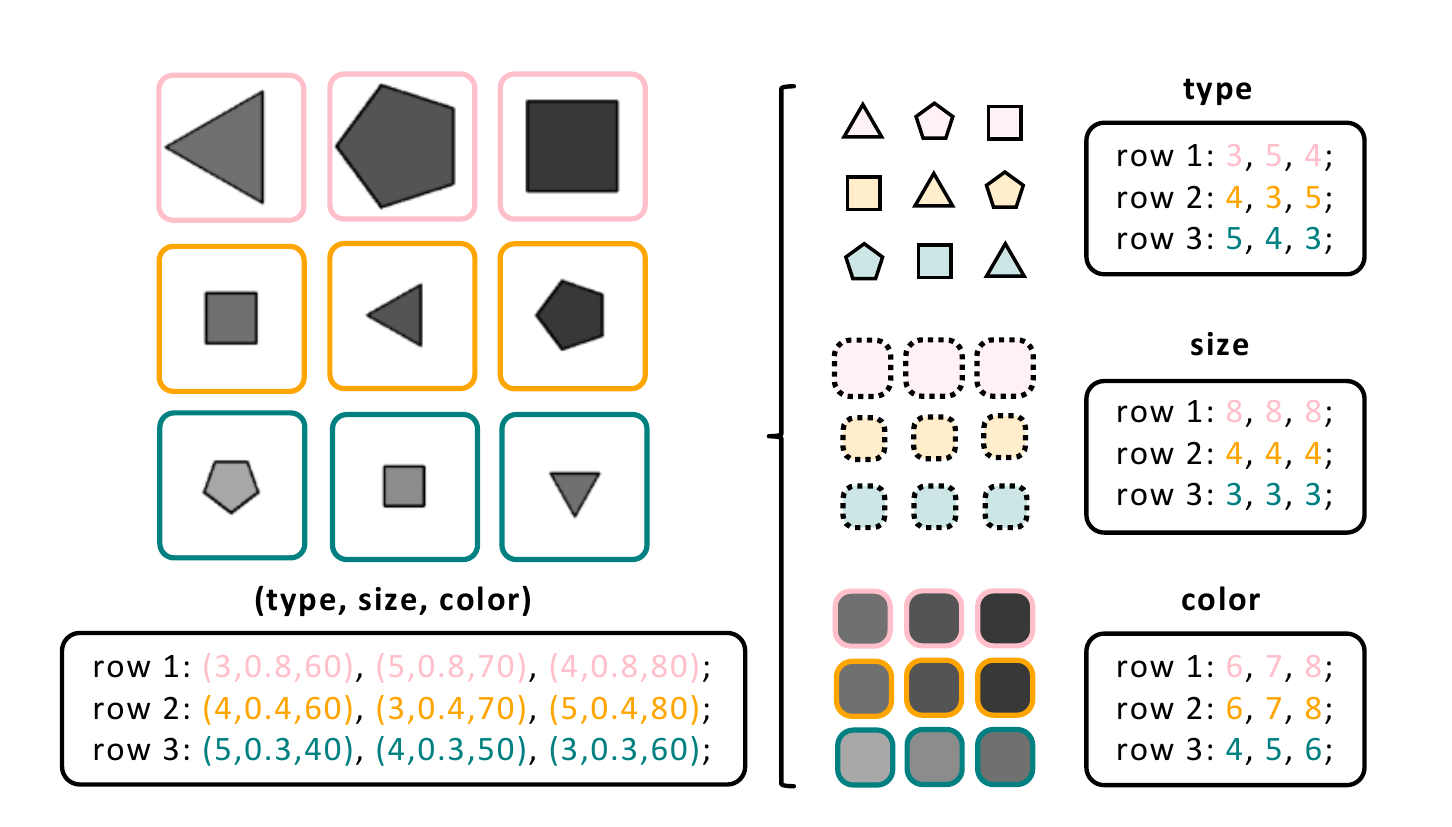}
    \caption{Example generated prompts for a complete RPM under entity attribute naming (left) and decomposition (right) abstractions in the \texttt{Center} sub-task.}
    \vspace{-10pt}
    \label{fig: center prompt example}
\end{figure}

\subsubsection{Entity-Level Abstractions}\label{sec:entity level abstractions}
As shown in Figure~\ref{fig: center prompt example}, 
we can abstract perceptual entity attributes into language by assigning them names, then generating prompts to represent the full RPM using these names.
As each of an entity's attributes is numerical by nature, we assign each attribute an ordinal numerical name; \texttt{type} is named by the number of sides of the associated shape (e.g., ``3'' for \textit{triangle}), \texttt{size} is named by a decimal representing its diameter, 
and \texttt{color} is named based on the darkness of the entity's shade.
As each of an entity's attributes is independent, i.e., a relation over one attribute has no connection to relations over other attributes, we can decompose the RPM task by these attributes into three separate sub-tasks with their own prompts.

\begin{figure}
    \centering
    \includegraphics[width=0.49\textwidth]{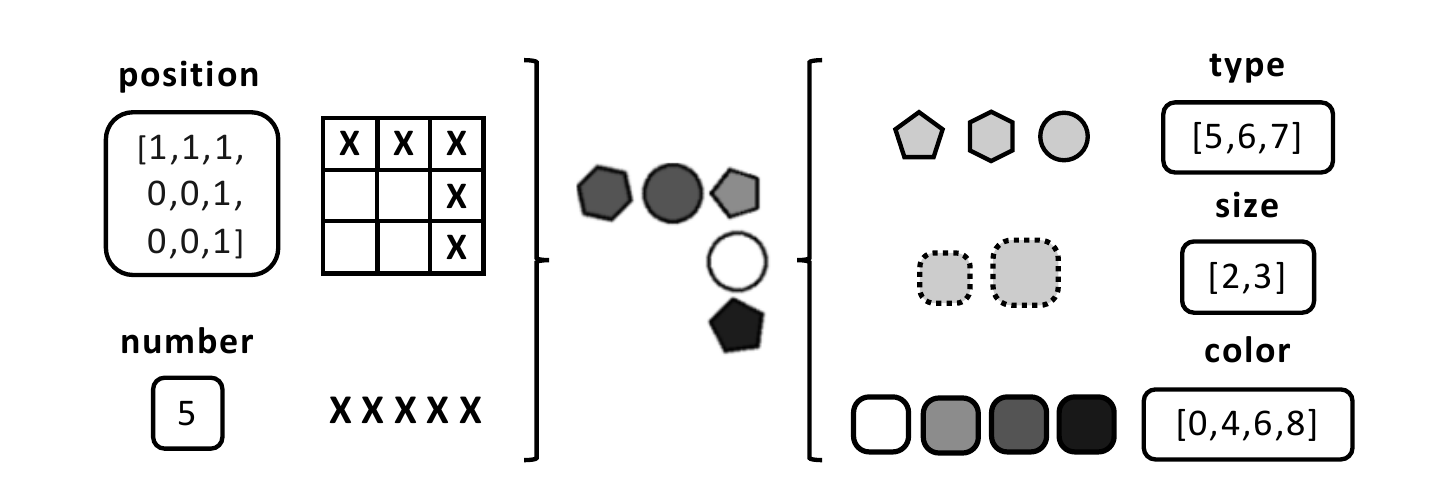}
    \caption{Example of generated entity layout encodings when abstracting \texttt{position} and \texttt{number}, and summarizing redundant entity attributes within the layout.}
    \vspace{-10pt}
    \label{fig: layout prompt example}
\end{figure}

\subsubsection{Layout-Level Abstractions}
As shown in Figure~\ref{fig: layout prompt example}, we next propose abstractions for layouts of entities (e.g., in grid-based sub-tasks of RAVEN).
First, the \texttt{number} attribute of a layout corresponds to the count of entities in it. Recognizing \texttt{number} requires implicitly counting entities within a layout, which may be difficult to disentangle from other attributes.
As such, we directly expose this attribute by extracting this count and encoding it in text. Since this layout attribute is independent from other attributes, we can again decompose the task and consider it separately from entity attributes.

The \texttt{position} attribute encodes even more complex information about a layout, and relations over it may move entities around within the layout. 
However, an occupancy map serves as a strong naming abstraction for \texttt{position} which omits distracting details of specific entities while exposing key information for detecting relations over it.
We generate the occupancy map as an array of text representing the occupancy of the layout, and decompose this from other attributes.
Notably, this abstraction provides a unique language description for each possible global configuration of entities within a layout, allowing the PLM to disentangle global and local patterns in the problem, a helpful capability of
humans \cite{ROBERTSON1991299}.\footnote{For example, we may recognize the grid of entities in Figure~\ref{fig: entity layout structure} to be in an ``L'' shape at the global level, while also recognizing that it is locally composed of triangles.} 

In RAVEN, relations are applied to specific attributes consistently across all entities in a layout. As our layout-level abstractions make explicit the key features of layouts, we no longer need to track entity-level attributes for specific entities within them. Specifically, rather than supply a PLM with a separate grid-like prompt for each entity-level attribute, we simply provide a list of unique attribute values. This reduces the 
complexity added by layouts of multiple entities.

\subsubsection{Structural Decomposition Abstractions}
In cases with multiple components in each item, we may find that prompts become long and complicated with earlier approaches.
Since each component's attributes and relations are independent, we can alternatively decompose the task by its components.
For each component, we can generate a prompt through entity attribute naming abstractions as shown in Figure~\ref{fig: center prompt example} (left), or we can apply the higher-level abstractions over entity and layout attributes shown in Figure~\ref{fig: layout prompt example}, thus decomposing each component's prompts into prompts for each attribute.
As this structural decomposition converts multi-component problems into several simpler single-component, single-attribute problems, the complexity added by multiple components is abstracted away.

\subsection{Problem Definition}\label{sec:formal definitions}
Formally, a complete RPM $M$ consists of 9 matrix items $m_{ij}$ where row and column $i,j\in\{1,2,3\}$.
As discussed in Section~\ref{sec:datasets}, an individual item $m_{ij}$ in the RAVEN dataset is formalized by high-level components consisting of layout-level attributes and entity-level attributes. 
Given all items in $M$ except for $m_{33}$, the task is to identify $m_{33}$ from a set $Y$ of $8$ choices by identifying abstract rules over the attributes within the first 2 rows of $M$, and selecting the candidate $m_{33}$ that correctly applies these rules in the third row.

\paragraph{Applying PLMs.} We apply PLMs to RAVEN in a zero-shot setting. In the absence of decomposition abstractions, we define $\mathbbm{L}$ as the mapping of a complete RPM to a text prompt. The PLM's choice for $m_{33}$ is given by
\[ \arg\max_{y\in Y}\frac{1}{|\mathbbm{L}|}\log\Pr\left(\mathbbm{L}\left(m_{11:32},y\right)\right)\] where $|\mathbbm{L}|$ denotes the number of tokens in the prompt. When decomposition is introduced, $\mathbbm{L}$ instead returns multiple prompts, and the (token-length normalized) log-probabilities of all sub-prompts are summed.\footnote{See Appendix~\ref{apx: example prompts} for examples of decomposing prompts.}

\section{Experimental Results}
Now, we can examine the impact each of these language-based abstractions has on the performance of transformer-based, autoregressive PLMs in relational reasoning on RAVEN. To further understand their impact with respect to model complexity, we evaluate a range of model sizes:\footnote{Results on additional model sizes in Appendix~\ref{apx: expanded results}.} OPT 125M, 1.3B, and 13B~\cite{zhang2022opt}, along with GPT-3~\cite{brown2020language}.\footnote{Specifically, we use the \texttt{text-davinci-002} variant of InstructGPT~\cite{ouyang2022training} through a Microsoft Azure OpenAI deployment.} Models are evaluated on a random subset of 500 testing examples from each sub-task of RAVEN.

After introducing some comparison approaches, we present the experimental results from our applied abstractions on PLMs' entity-level, layout-level, and component-level relational reasoning. Afterward, we dive deeper with an analysis on how both our abstractions and in-context learning contribute to model performance.

\subsection{Comparison Approaches}\label{sec:simple baselines}
{To contextualize our findings,} we provide results from the human study in \citet{zhang2019raven}, as well as two supervised baselines from prior work.\footnote{{Since our approach is not evaluated on the exact same subset of RAVEN data, these results from prior work are not directly comparable, but can be helpful reference points.}} Additionally, to specifically evaluate the advantage of the way we mapped the RPM task into language, we include two simpler abstraction methods that encode task information less explicitly.

\paragraph{Supervised baselines.}
{While our goal is not to achieve the state of the art on RAVEN, we include results from two state-of-the-art supervised baselines for reference. Specifically, we select the two approaches with the top mean accuracy on RAVEN, as outlined in the survey by \citet{malkinski2022deep}: Rel-AIR~\cite{spratley2020closer} and CoPINet + ACL~\cite{NEURIPS2020_c39e1a03}. Rel-AIR combines a simple vision model with an unsupervised scene decomposition module, enabling more generalizable reasoning over entities in RAVEN. CoPINet + ACL applies an analogy-centric contrastive learning paradigm to CoPINet~\cite{NEURIPS2019_6766aa27}, a prior architecture proposed for perceptual inference trained through contrastive learning. Both baselines have been trained on thousands of examples from the RAVEN dataset, and incorporate task-specific inductive biases in their architecture. Meanwhile, we evaluate PLMs on RAVEN in a zero-shot setting with no supervised learning.}

\begin{figure}
    \centering



    \includegraphics[width=0.51\textwidth]{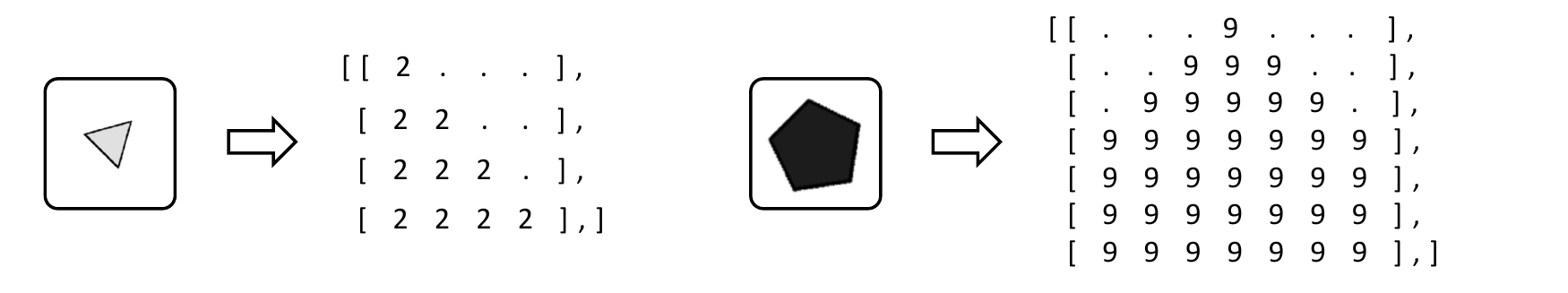}
    \caption{Quasi-image abstractions for a triangle and pentagon of different \texttt{size} and \texttt{color}.}
    \label{fig:quasi pixel}
\end{figure}

\paragraph{Quasi-image abstraction.}
To evaluate the helpfulness of naming abstractions over entity attributes, we should compare to an approach that does not have such abstraction. However, some mapping from the visual features of the RPM task into langauge is needed in order for a PLM to interface with it. 
While the limited context window of PLMs restricts us from incorporating raw pixels directly into our prompts, PLMs have recently been demonstrated to capture spatial patterns in similar inputs: text-based matrices \cite{patel2021mapping}. 
As such, we propose a \textit{quasi-image} abstraction which converts the visual RPM task into a matrix of ASCII characters. As shown in Figure~\ref{fig:quasi pixel}, an entity's \texttt{type} can be expressed through a matrix of characters; \texttt{size} can be expressed through the height and width of the matrix; and \texttt{color} can be expressed through the actual characters making up the matrix. By converting instances of RAVEN's \texttt{Center} sub-task into this pixel-like form, we have a lower-level abstraction of the task's visual features that can be compared to the higher-level abstraction of naming entity attributes.

\paragraph{Random naming abstraction.}
We would also like to understand the advantage of the specific names we chose for entity attributes compared to other possible choices.
As such, we propose a second baseline where, instead of using ordinal labels to describe entities' \texttt{type}, \texttt{size}, and \texttt{color}, we choose random words from a large corpus. This removes numerical dependencies that may be utilized to recognize some relations, and can help us understand whether PLMs take advantage of this information when it is available.

\subsection{Entity-Level Reasoning}
We first evaluate PLMs under our lowest level abstractions over entity attributes. To isolate the improvements from such abstraction, we focus on the \texttt{Center} sub-task of RAVEN which only includes a single entity per item in the RPM, and thus only tests understanding of relations over entity attributes. The results are shown in Figure~\ref{fig:center results}.

\definecolor{darkgray}{rgb}{0.45, 0.45, 0.45}
\definecolor{lightgray}{rgb}{0.62, 0.62, 0.62}

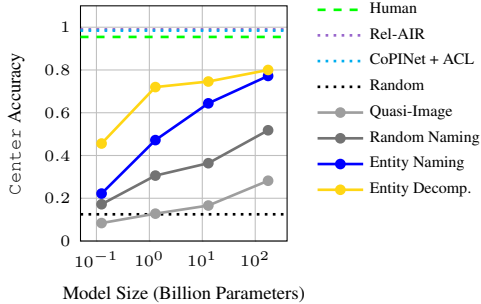
\begin{figure}
\centering
\vspace{-10pt}
\begin{tikzpicture}
\begin{axis}
[xlabel={\scriptsize Model Size (Billion Parameters)}, ylabel={\scriptsize\texttt{Center} Accuracy}, width=4.3cm, height=4.7cm, xmode=log, legend style={at={(2,0.6)},anchor=east,draw=none}, legend columns=1, legend cell align=left, xtick={0.1,1,10,100}, ytick={0,0.2,0.4,0.6,0.8,1}, minor tick style={draw=none}, xtick pos=bottom, xtick align=outside, ytick pos=left, ytick align=outside, major tick length=0, xmin=0.05, xmax=400, ymin=0, ymax=1.1, compat=1.18, xmajorgrids=true, ymajorgrids=true]

\addplot[color=green,dashed,line width=1pt] coordinates {(0.05, 0.9545) (400, 0.9545)};
\addlegendentry{\tiny Human}

\addplot[color=purple,dotted,line width=1pt] coordinates {(0.05, 0.990) (400, 0.990)};
\addlegendentry{\tiny Rel-AIR}

\addplot[color=cyan,dotted,line width=1pt] coordinates {(0.05, 0.984) (400, 0.984)};
\addlegendentry{\tiny CoPINet + ACL}

\addplot[color=black,dotted,line width=1pt] coordinates {(0.05, 0.125) (400, 0.125)};
\addlegendentry{\tiny Random}

\addplot[mark=*,mark size=1.5pt,mark options={color=lightgray,fill=lightgray},color=lightgray,line width=1pt] coordinates {(0.125, 0.084) (1.3, 0.128) (13, 0.166) (175, 0.282)};
\addlegendentry{\tiny Quasi-Image}

\addplot[mark=*,mark size=1.5pt,mark options={color=darkgray,fill=darkgray},color=darkgray,line width=1pt] coordinates {(0.125, 0.172) (1.3, 0.306) (13, 0.364) (175, 0.518)};
\addlegendentry{\tiny Random Naming}

\addplot[mark=*,mark size=1.5pt,color=blue,line width=1pt] coordinates {(0.125, 0.222) (1.3, 0.472) (13, 0.644) (175, 0.772)};
\addlegendentry{\tiny Entity Naming}

\addplot[mark=*,mark size=1.5pt,color=gold,line width=1pt] coordinates {(0.125, 0.456) (1.3, 0.72)  (13, 0.746) (175, 0.8)};
\addlegendentry{\tiny Entity Decomp.}

\end{axis}
\end{tikzpicture}

\caption{Results on the RAVEN \texttt{Center} sub-task under entity abstractions, compared to na{\"i}ve and supervised baselines described in Section~\ref{sec:simple baselines}, and humans.}
\vspace{-15pt}
\label{fig:center results}
\end{figure}

\paragraph{Impact of naming.}
Under the simplest abstraction of naming the entity-level attributes, we see impressive zero-shot accuracies that monotonically increase with model size up to 77.2\% from GPT-3 175B on \texttt{Center}, nearing human performance. Further, we find that our choice to map attributes into numerical symbols is consistently advantageous over the quasi-image and random-naming abstractions, which reach respective accuracies up to 28.2\% and 51.8\%. 
Meanwhile, we find that as model size increases, our ordinal naming approach outperforms the random naming baseline more and more, up to over 20\% in larger model sizes. This suggests that PLMs of larger size can better capture and take advantage of implicit numerical relations in their vocabulary.

\paragraph{Impact of decomposition.}
When applying decomposition over entity attributes, we observe further improvement of 2.8\% accuracy in GPT-3 175B. Interestingly, we see a much sharper improvement from this abstraction in smaller models, with OPT 125M's accuracy doubling from 22.2\% to 45.6\%, and OPT 1.3B's accuracy rising from 47.2\% to 72.0\%. This may suggest that PLMs have a limited working memory which is related to the number of learned parameters in them. Large PLMs are more capable to handle complex reasoning tasks because of this, while smaller PLMs benefit from decomposing tasks into more manageable parts.

\subsection{Layout-Level Reasoning}
In Figure~\ref{fig: layout results}, we evaluate PLMs' capability to capture relations over layout attributes under our abstractions introduced in the \texttt{2x2Grid} and \texttt{3x3Grid} sub-tasks. 
Without any decomposition abstraction, model performance reaches up to 78.0\% and 86.4\% accuracy respectively on \texttt{2x2Grid} and \texttt{3x3Grid}. When adding naming for layout-level attributes and decomposing all attributes into separate prompts, we see further improvements across the board, with accuracies reaching 87.8\% on \texttt{2x2Grid} and 93.2\% on \texttt{3x3Grid}. The PLM exceeds human performance on both sub-tasks, despite them being arguably some of the most complex tasks in RAVEN, with the latter comprised of more entities than any other sub-task.
This suggests that our strong layout-level abstractions enable the PLM to tease apart the numerous attributes in grids of entities and capture obscure patterns, whereas humans may struggle with this as the task becomes more complex.

\begin{figure}[h]
\centering
\begin{tikzpicture}
\begin{groupplot}[group style={group size=2 by 1, horizontal sep=1cm}, width=0.2\textwidth, legend style={draw=none, anchor=north, legend columns=3}, legend to name=commonlegend, legend cell align=left, xlabel={}, xmode=log, xtick={0.1,1,10,100}, ytick={0,0.2,0.4,0.6,0.8,1}, minor tick style={draw=none}, xtick pos=bottom, xtick align=outside, ytick pos=left, ytick align=outside, major tick length=0, xmin=0.05, xmax=400, ymin=0, ymax=1.1, compat=1.18, xmajorgrids=true, ymajorgrids=true]

\nextgroupplot[ylabel={\scriptsize\texttt{2x2Grid} Accuracy}, ylabel style={yshift=-5pt}, xlabel={\scriptsize Model Size (Billion Parameters)}, xlabel style={at={(1,-0.2)}}, width=4.3cm, height=4.7cm]

\addplot[color=green,dashed,line width=1pt] coordinates {(0.05, 0.818) (400, 0.818)};
\addlegendentry{\tiny Human}

\addplot[color=purple,dotted,line width=1pt] coordinates {(0.05, 0.924) (400, 0.924)};
\addlegendentry{\tiny Rel-AIR}

\addplot[color=cyan,dotted,line width=1pt] coordinates {(0.05, 0.810) (400, 0.810)};
\addlegendentry{\tiny CoPINet + ACL}

\addplot[color=black,dotted,line width=1pt] coordinates {(0.05, 0.125) (400, 0.125)};
\addlegendentry{\tiny Random}

\addplot[mark=*,mark size=1.5pt,color=blue,line width=1pt] coordinates {(0.125, 0.42) (1.3, 0.584) (13, 0.61) (175, 0.78)};
\addlegendentry{\tiny Entity Naming}

\addplot[mark=*,mark size=1.5pt,color=gold,line width=1pt] coordinates {(0.125, 0.62) (1.3, 0.714) (13, 0.794) (175, 0.878)};
\addlegendentry{\tiny Entity\& Layout Decomp.}

\nextgroupplot[ylabel={\scriptsize\texttt{3x3Grid} Accuracy}, ylabel style={yshift=-5pt}, width=4.3cm, height=4.7cm]

\addplot[color=green,dashed,line width=1pt] coordinates {(0.05, 0.796) (400, 0.796)};
\addlegendentry{\tiny\color{black}Human}

\addplot[color=purple,dotted,line width=1pt] coordinates {(0.05, 0.871) (400, 0.871)};
\addlegendentry{\tiny\color{black}Rel-AIR}

\addplot[color=cyan,dotted,line width=1pt] coordinates {(0.05, 0.840) (400, 0.840)};
\addlegendentry{\tiny\color{black}CoPINet + ACL}

\addplot[color=black,dotted,line width=1pt] coordinates {(0.05, 0.125) (400, 0.125)};
\addlegendentry{\tiny\color{black}Random}

\addplot[mark=*,mark size=1.5pt,color=blue,line width=1pt] coordinates {(0.125, 0.606) (1.3, 0.71) (13, 0.754) (175, 0.864)};
\addlegendentry{\tiny\color{black}Entity Naming}

\addplot[mark=*,mark size=1.5pt,color=gold,line width=1pt] coordinates {(0.125, 0.724) (1.3, 0.794) (13, 0.83) (175, 0.932)};
\addlegendentry{\tiny\color{black}Entity \& Layout Decomp.}

\end{groupplot}
\node at ($(group c1r1.center)+(1.5,2cm)$) {\ref{commonlegend}};

\end{tikzpicture}
\caption{Results on grid-based sub-tasks of RAVEN without and with decomposition abstractions. Compared to humans and supervised baselines.}
\vspace{-10pt}
\label{fig: layout results}
\end{figure}
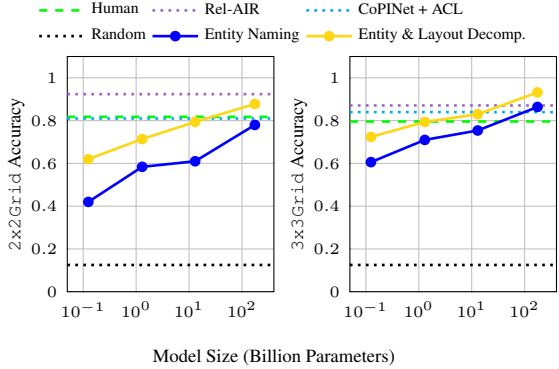

\begin{figure*}[h]
\centering
\begin{tikzpicture}
\begin{groupplot}
[group style={group size=4 by 1, horizontal sep=1cm}, width=0.2\textwidth, legend style={draw=none, anchor=north, legend columns=4}, legend to name=commonlegend, legend cell align=left, xlabel={}, xmode=log, xtick={0.1,1,10,100}, ytick={0,0.2,0.4,0.6,0.8,1}, minor tick style={draw=none}, xtick pos=bottom, xtick align=outside, ytick pos=left, ytick align=outside, major tick length=0, xmin=0.05, xmax=400, ymin=0, ymax=1.1, compat=1.18, xmajorgrids=true, ymajorgrids=true]

\nextgroupplot[ylabel={\scriptsize\texttt{L-R} Accuracy}, ylabel style={yshift=-5pt}, xlabel={\scriptsize Model Size (Billion Parameters)}, xlabel style={at={(2.5,-0.2)}}, width=4.3cm, height=4.7cm]

\addplot[color=green,dashed,line width=1pt] coordinates {(0.05, 0.8636) (400, 0.8636)};
\addlegendentry{\tiny\color{black}Human}

\addplot[color=purple,dotted,line width=1pt] coordinates {(0.05, 0.987) (400, 0.987)};
\addlegendentry{\tiny\color{black}Rel-AIR}

\addplot[color=cyan,dotted,line width=1pt] coordinates {(0.05, 0.997) (400, 0.997)};
\addlegendentry{\tiny\color{black}CoPINet + ACL}

\addplot[color=black,dotted,line width=1pt] coordinates {(0.05, 0.125) (400, 0.125)};
\addlegendentry{\tiny\color{black}Random}

\addplot[mark=*,mark size=1.5pt,color=blue,line width=1pt] coordinates {(0.125, 0.076) (1.3, 0.146) (13, 0.22) (175, 0.542)};
\addlegendentry{\tiny\color{black}Attr. Naming}

\addplot[mark=*,mark size=1.5pt,color=red,line width=1pt] coordinates {(0.125, 0.136) (1.3, 0.41) (13, 0.566) (175, 0.738)};
\addlegendentry{\tiny\color{black}Comp. Decomp.}

\addplot[mark=*,mark size=1.5pt,color=gold,line width=1pt] coordinates {(0.125, 0.378) (1.3, 0.672) (13, 0.71) (175, 0.776)};
\addlegendentry{\tiny\color{black}Comp. \& Attr. Decomp.}

\nextgroupplot[ylabel={\scriptsize\texttt{U-D} Accuracy}, ylabel style={yshift=-5pt}, width=4.3cm, height=4.7cm]

\addplot[color=green,dashed,line width=1pt] coordinates {(0.05, 0.8181) (400, 0.8181)};
\addlegendentry{\tiny\color{black}Human}

\addplot[color=purple,dotted,line width=1pt] coordinates {(0.05, 0.979) (400, 0.979)};
\addlegendentry{\tiny\color{black}Rel-AIR}

\addplot[color=cyan,dotted,line width=1pt] coordinates {(0.05, 0.998) (400, 0.998)};
\addlegendentry{\tiny\color{black}CoPINet + ACL}

\addplot[color=black,dotted,line width=1pt] coordinates {(0.05, 0.125) (400, 0.125)};
\addlegendentry{\tiny\color{black}Random}

\addplot[mark=*,mark size=1.5pt,color=blue,line width=1pt] coordinates {(0.125, 0.098) (1.3, 0.158) (13, 0.268) (175, 0.536)};
\addlegendentry{\tiny\color{black}Attr. Naming}

\addplot[mark=*,mark size=1.5pt,color=red,line width=1pt] coordinates {(0.125, 0.154) (1.3, 0.426) (13, 0.602) (175, 0.732)};
\addlegendentry{\tiny\color{black}Comp. Decomp.}

\addplot[mark=*,mark size=1.5pt,color=gold,line width=1pt] coordinates {(0.125, 0.408) (1.3, 0.68) (13, 0.702) (175, 0.78)};
\addlegendentry{\tiny\color{black}Comp. \& Attr. Decomp.}

\nextgroupplot[ylabel={\scriptsize\texttt{O-IC} Accuracy}, ylabel style={yshift=-5pt}, width=4.3cm, height=4.7cm]

\addplot[color=green,dashed,line width=1pt] coordinates {(0.05, 0.8636) (400, 0.8636)};
\addlegendentry{\tiny\color{black}Human}

\addplot[color=purple,dotted,line width=1pt] coordinates {(0.05, 0.980) (400, 0.980)};
\addlegendentry{\tiny\color{black}Rel-AIR}

\addplot[color=cyan,dotted,line width=1pt] coordinates {(0.05, 0.994) (400, 0.994)};
\addlegendentry{\tiny\color{black}CoPINet + ACL}

\addplot[color=black,dotted,line width=1pt] coordinates {(0.05, 0.125) (400, 0.125)};
\addlegendentry{\tiny\color{black}Random}

\addplot[mark=*,mark size=1.5pt,color=blue,line width=1pt] coordinates {(0.125, 0.122) (1.3, 0.2) (13, 0.358) (175, 0.648)};
\addlegendentry{\tiny\color{black}Attr. Naming}

\addplot[mark=*,mark size=1.5pt,color=red,line width=1pt] coordinates {(0.125, 0.162) (1.3, 0.434) (13, 0.586) (175, 0.78)};
\addlegendentry{\tiny\color{black}Comp. Decomp.}

\addplot[mark=*,mark size=1.5pt,color=gold,line width=1pt] coordinates {(0.125, 0.374) (1.3, 0.744) (13, 0.77) (175, 0.828)};
\addlegendentry{\tiny\color{black}Comp. \& Attr. Decomp.}







\nextgroupplot[ylabel={\scriptsize\texttt{O-IG} Accuracy}, ylabel style={yshift=-5pt}, width=4.3cm, height=4.7cm]

\addplot[color=green,dashed,line width=1pt] coordinates {(0.05, 0.818) (400, 0.818)};
\addlegendentry{\tiny\color{black}Human}

\addplot[color=purple,dotted,line width=1pt] coordinates {(0.05, 0.853) (400, 0.853)};
\addlegendentry{\tiny\color{black}Rel-AIR}

\addplot[color=cyan,dotted,line width=1pt] coordinates {(0.05, 0.939) (400, 0.939)};
\addlegendentry{\tiny\color{black}CoPINet + ACL}

\addplot[color=black,dotted,line width=1pt] coordinates {(0.05, 0.125) (400, 0.125)};
\addlegendentry{\tiny\color{black}Random}

\addplot[mark=*,mark size=1.5pt,color=blue,line width=1pt] coordinates {(0.125, 0.194) (1.3, 0.322) (13, 0.452) (175, 0.748)};
\addlegendentry{\tiny\color{black}Attr. Naming}

\addplot[mark=*,mark size=1.5pt,color=red,line width=1pt] coordinates {(0.125, 0.222) (1.3, 0.494) (13, 0.576) (175, 0.84)};
\addlegendentry{\tiny\color{black}Comp. Decomp.}

\addplot[mark=*,mark size=1.5pt,color=gold,line width=1pt] coordinates {(0.125, 0.52) (1.3, 0.744) (13, 0.84) (175, 0.926)};
\addlegendentry{\tiny\color{black}Comp. \& Attr. Decomp.}

\end{groupplot}
\node at ($(group c1r1.center)+(5.4,2cm)$) {\ref{commonlegend}};

\end{tikzpicture}
\caption{PLM accuracy on multi-component RAVEN sub-tasks with attribute naming only, component decomposition, and full component and attribute decomposition, compared to supervised baselines and humans.}
\vspace{-5pt}
\label{fig: component results}
\end{figure*}
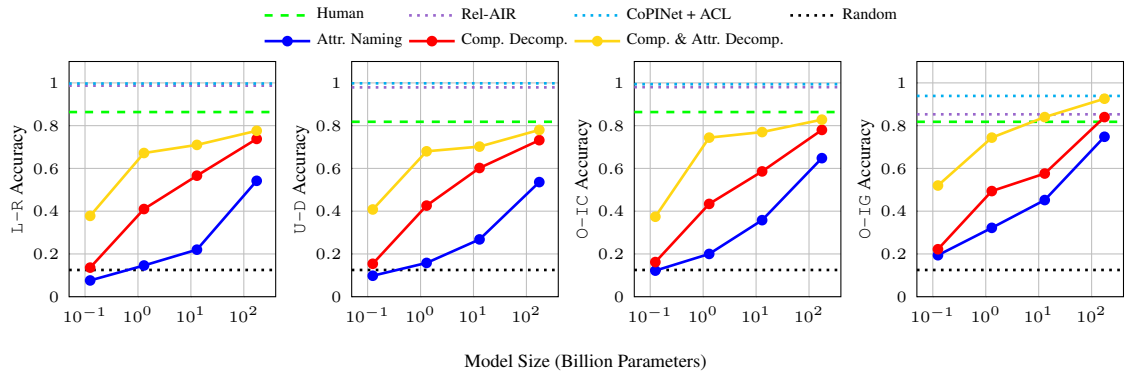

\subsection{Component-Level Reasoning}
Lastly, we apply our structural decomposition-based abstractions on RAVEN sub-tasks which have multiple components, i.e., \texttt{L-R}, \texttt{U-D}, \texttt{O-IC}, and \texttt{O-IG}. The results are shown in Figure~\ref{fig: component results}. 
First, just decomposing the task by its components improves the maximum accuracy on each task on average by about 20\%. Additionally decomposing each component by its entity and layout attributes brings further gains, with GPT-3 175B reaching up to 77.6\%, 78.0\%, 82.8\%, and 92.6\% on \texttt{L-R}, \texttt{U-D}, \texttt{O-IC}, and \texttt{O-IG} respectively, and exceeding humans and nearing supervised baselines on the latter.
The performance gain from this decomposition is again even more pronounced for smaller PLMs. 
Most significantly, OPT 1.3B improves from 20-30\% accuracy to over 70\% accuracy, nearing human performance. This demonstrates that not only is GPT-3 capable of very complex analogical reasoning tasks, but even PLMs less than 100 times its size can perform quite well here with the proper abstractions.

\subsection{Fine-Grained Analysis}\label{sec:analysis}
Finally, we analyze how model performance varies across different attributes and relations, as we introduce distracting attributes, and as we introduce rows into the matrix. In our analysis, we compare three representative levels of abstraction: \textit{entity attribute naming only} (no decomposition into multiple prompts), \textit{decomposition of components}, and full \textit{decomposition of entity and layout attributes and components}.

\subsubsection{Analysis of Attributes and Relations}
\begin{figure}
   \centering    
   \includegraphics[width=0.48\textwidth]{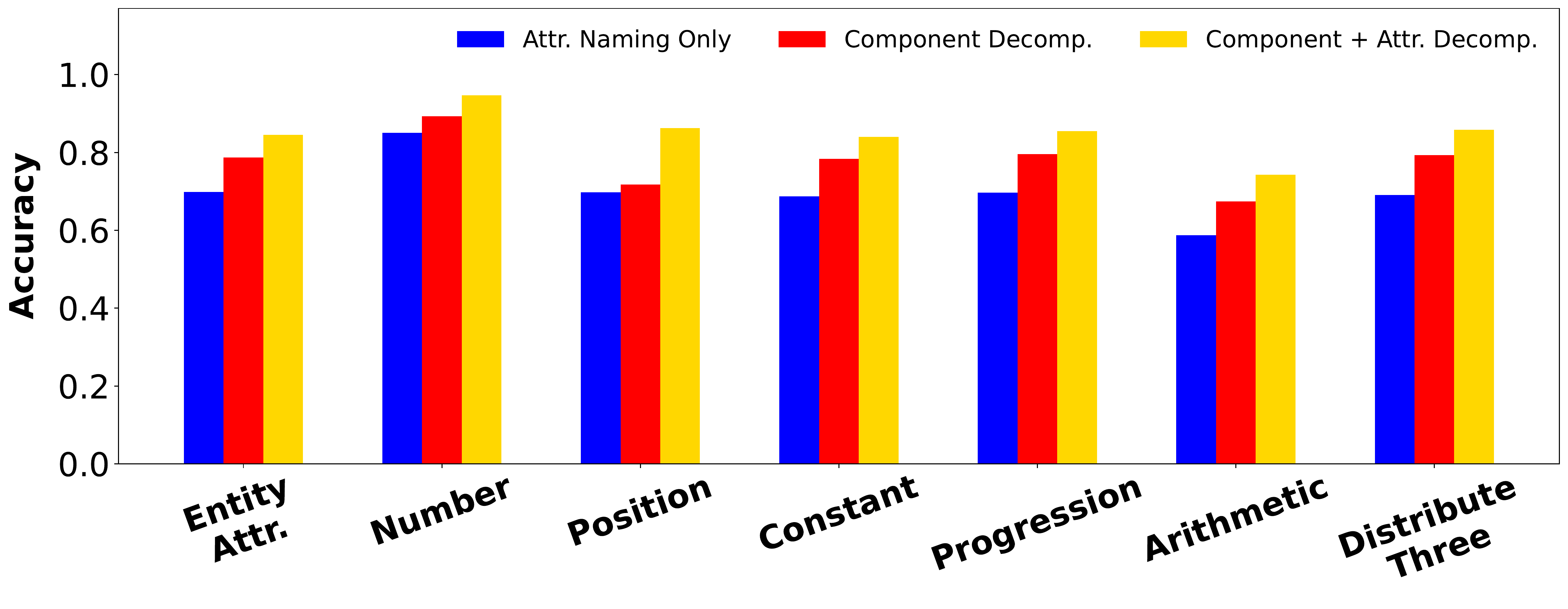}
   \vspace{-20pt}
   \caption{Comparison of accuracy on examples from all sub-tasks, broken down by the types of attributes and relations they require capturing.}
   \vspace{-5pt}
   \label{fig: abstractions analysis}
\end{figure}

We measure the impact of abstractions in capturing each attribute and relation in RAVEN. In Figure~\ref{fig: abstractions analysis}, we present GPT-3 175B's accuracy over each attribute and relation.
We find that \texttt{number} is the best captured attribute even without any decomposition abstractions, while the model struggles with \texttt{position} until we introduce decomposition of attributes, suggesting the occupancy map encoding used here indeed helped capture it.
Meanwhile, \texttt{Arithmetic} is the most difficult relation, with consistently lower accuracy than other relations.

\subsubsection{Robustness to Distracting Attributes}\label{sec: distractor analysis}
{Since our mappings from RAVEN attributes into language provide the key features over which relations occur, we may wonder how robust PLMs are to distracting or unimportant attributes. In fact, the RAVEN dataset includes one noise attribute that we excluded from our mapping to avoid unnecessarily increasing prompt lengths: \texttt{orientation}, i.e., the rotation of entities in the RPM. To begin exploring this issue, we incorporate \texttt{orientation} into the problem as a fourth entity-level attribute in addition to \texttt{type}, \texttt{size}, and \texttt{color}. For the best model (i.e., GPT-3) on the \texttt{Center} sub-task, we compare two possible injections of \texttt{orientation} values: using the values provided in RAVEN (which are mostly constant within each matrix row), and randomly selected values (which could be more distracting).}

{As shown in Table~\ref{tab: rotation analysis}, compared to GPT-3's \texttt{Center} accuracies of 77.2\% and 80.0\% with respective naming and decomposition abstractions, the injection of \texttt{orientation} as a distraction feature does not degrade the model performance much, achieving accuracies of 76.0\% and 80.0\% when using values from RAVEN, and 72.6\% and 77.8\% when using random values. This shows that PLMs exhibit some robustness to distracting attributes in language context, and have the capability to ignore them in analogical reasoning. Future work may consider more in-depth analysis to discover the extent of model robustness to distraction features, and how it varies by model complexity.}

\begin{table}
    \centering
    \footnotesize

    \setlength\tabcolsep{5pt}
    \begin{tabular}{ccccc}\toprule
        \textbf{Distractor Values} & \textbf{Naming} & \textbf{Decomposition} \\
         \cmidrule(lr){1-1} \cmidrule(lr){2-3}
        RAVEN & 76.0\% & 80.0\% \\
        Random & 72.6\% & 77.8\% \\
         \bottomrule
    \end{tabular}

    \normalsize
    \caption{GPT-3 accuracy on \texttt{Center} sub-task with distracting \texttt{orientation} attribute in language prompts, under the naming and decomposition abstractions. \texttt{orientation} values are taken directly from RAVEN or randomly selected.}
    \vspace{-10pt}

    \label{tab: rotation analysis}
\end{table}

\subsubsection{In-Context Learning Over Rows}\label{sec: prior analysis}
By design, RPM tasks are meant to require minimal background knowledge. They should be impossible to solve without the first two rows of the matrix, which provide essential context to complete the third row of the matrix. To understand whether PLMs capture relations specifically from in-context learning over the first two rows of the matrix (as opposed to using prior knowledge from pre-training), we measure the model performance as we introduce rows to the matrices.

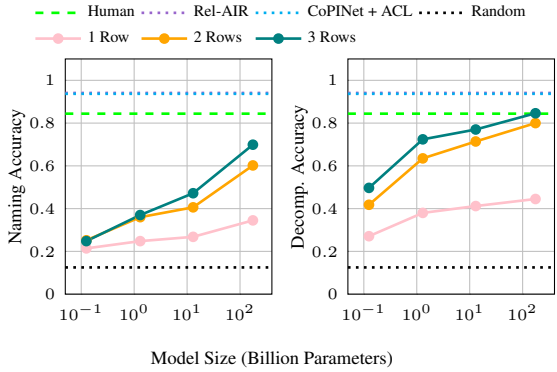
\begin{figure}[h]
\centering
\vspace{-7pt}
\begin{tikzpicture}
\begin{groupplot}[group style={group size=2 by 1, horizontal sep=1cm}, width=0.2\textwidth, legend style={draw=none, anchor=north, legend columns=4}, legend to name=commonlegend, legend cell align=left, xlabel={}, xmode=log, xtick={0.1,1,10,100}, ytick={0,0.2,0.4,0.6,0.8,1}, minor tick style={draw=none}, xtick pos=bottom, xtick align=outside, ytick pos=left, ytick align=outside, major tick length=0, xmin=0.05, xmax=400, ymin=0, ymax=1.1, compat=1.18, xmajorgrids=true, ymajorgrids=true]

\nextgroupplot[ylabel={\scriptsize Naming Accuracy}, ylabel style={yshift=-6pt}, xlabel={\scriptsize Model Size (Billion Parameters)}, xlabel style={at={(1,-0.2)}}, width=4.3cm, height=4.7cm]

\addplot[color=green,dashed,line width=1pt] coordinates {(0.05, 0.8441) (400, 0.8441)};
\addlegendentry{\tiny Human}

\addplot[color=purple,dotted,line width=1pt] coordinates {(0.05, 0.941) (400, 0.941)};
\addlegendentry{\tiny\color{black}Rel-AIR}

\addplot[color=cyan,dotted,line width=1pt] coordinates {(0.05, 0.937) (400, 0.937)};
\addlegendentry{\tiny\color{black}CoPINet + ACL}

\addplot[color=black,dotted,line width=1pt] coordinates {(0.05, 0.125) (400, 0.125)};
\addlegendentry{\tiny Random}

\addplot[mark=*,mark size=1.5pt,color=pink,line width=1pt] coordinates {(0.125, 0.214) (1.3, 0.248) (13, 0.268) (175, 0.345)};
\addlegendentry{\tiny\color{black}1 Row}

\addplot[mark=*,mark size=1.5pt,color=orange,line width=1pt] coordinates {(0.125, 0.252) (1.3, 0.36) (13, 0.406) (175, 0.602)};
\addlegendentry{\tiny\color{black}2 Rows}

\addplot[mark=*,mark size=1.5pt,color=teal,line width=1pt] coordinates {(0.125, 0.248) (1.3, 0.37) (13, 0.472) (175, 0.699)};
\addlegendentry{\tiny\color{black}3 Rows}

\nextgroupplot[ylabel={\scriptsize Decomp. Accuracy}, ylabel style={yshift=-6pt}, width=4.3cm, height=4.7cm]

\addplot[color=green,dashed,line width=1pt] coordinates {(0.05, 0.8441) (400, 0.8441)};
\addlegendentry{\tiny\color{black}Human}

\addplot[color=purple,dotted,line width=1pt] coordinates {(0.05, 0.941) (400, 0.941)};
\addlegendentry{\tiny\color{black}Rel-AIR}

\addplot[color=cyan,dotted,line width=1pt] coordinates {(0.05, 0.937) (400, 0.937)};
\addlegendentry{\tiny\color{black}CoPINet + ACL}

\addplot[color=black,dotted,line width=1pt] coordinates {(0.05, 0.125) (400, 0.125)};
\addlegendentry{\tiny\color{black}Random}

\addplot[mark=*,mark size=1.5pt,color=pink,line width=1pt] coordinates {(0.125, 0.271) (1.3, 0.38) (13, 0.412) (175, 0.445)};
\addlegendentry{\tiny\color{black}1 Row}

\addplot[mark=*,mark size=1.5pt,color=orange,line width=1pt] coordinates {(0.125, 0.418) (1.3, 0.635) (13, 0.714) (175, 0.8)};
\addlegendentry{\tiny\color{black}2 Rows}

\addplot[mark=*,mark size=1.5pt,color=teal,line width=1pt] coordinates {(0.125, 0.497) (1.3, 0.724) (13, 0.77) (175, 0.846)};
\addlegendentry{\tiny\color{black}3 Rows}
\end{groupplot}
\node at ($(group c1r1.center)+(1.5,2cm)$) {\ref{commonlegend}};
\end{tikzpicture}
\vspace{-5pt}
\caption{Macro average accuracy over all RAVEN sub-tasks as we introduce rows to the matrix during in-context learning, under naming abstractions only (left) and all naming and decomposition abstractions (right). In 1 Row, we include only the incomplete third row.}
\label{fig: prior graph}
\end{figure}

As shown in Figure~\ref{fig: prior graph}, the average model performance increases across all sizes and abstractions as rows are added to the matrix.
This suggests that in-context learning indeed contributes significantly to performance, even for smaller models. Larger model sizes see the most significant improvements, suggesting that larger PLMs are stronger in-context learners than smaller ones. Further, larger PLMs can achieve nearly the same accuracy with only two rows of the matrix provided rather compared to having all three, suggesting that they pick up the task quite quickly from in-context learning.

We also observe that in many cases, models achieve accuracies above chance (12.5\% accuracy) without being provided any complete rows of the matrix (only the third, incomplete row). This may suggest the PLM has a useful prior for this problem, despite it being a visual problem and thus impossible to observe directly in pre-training. This raises questions about the objectivity of RAVEN and possibly the RPM task.\footnote{In Appendix~\ref{apx: iraven results}, we further explore this hypothesis on the Impartial-RAVEN dataset~\cite{hu2021stratified} that removes some superficial correlations in matrix completion choices, and still see comparable results.}
Further, when decomposition abstractions are applied, models achieve higher accuracies than when not, suggesting that decomposition encodes some of this prior knowledge for the task.
In Table~\ref{tab: prior table subtasks}, we take a closer look at GPT-3 175B's performance within sub-tasks. Surprisingly, we find the highest accuracies on the grid-based sub-tasks, despite them being the most difficult tasks for humans. 

This motivates future work to compare human and PLM performance on ablated analogy-making tasks like these to further evaluate their objectiveness and identify commonalities. Future work in AI and analogy may also consider building diagnostic datasets to tease apart attribute and relation types to better understand how they contribute to model performance and identify areas for improvement.

\begin{table}
    \centering
    \footnotesize

    \setlength\tabcolsep{5pt}
    \begin{tabular}{ccccc}\toprule
        \textbf{Sub-Task} & \textbf{1 Row} & \textbf{2 Rows} & \textbf{3 Rows} & \textbf{Human} \\
         \cmidrule(lr){1-1} \cmidrule(lr){2-4} \cmidrule(lr){4-4} \cmidrule(lr){5-5}
            
         \texttt{Center} & 36.8\% & 69.2\% & 77.2\% & 95.6\% \\
         \texttt{2x2Grid} & 54.0\% & 71.0\% & 78.0\% & 81.8\% \\
         \texttt{3x3Grid} & \textbf{73.0\%} & \textbf{85.2\%} & \textbf{86.4\%} & 79.6\% \\
         \texttt{L-R} & 14.0\% & 38.2\% & 54.2\% & 86.4\% \\
         \texttt{U-D} & 12.4\% & 42.0\% & 53.6\% & 81.8\% \\
         \texttt{O-IC} & 19.6\% & 53.6\% & 64.8\% & 86.4\% \\
         \texttt{O-IG} & 32.0\% & 62.2\% & 74.8\% & 81.8\% \\
         \bottomrule
    \end{tabular}

    \normalsize
    \caption{GPT-3 accuracy on RAVEN sub-tasks as rows are added to the RPM, under only naming abstractions.}
    \vspace{-10pt}

    \label{tab: prior table subtasks}
\end{table}

\begin{figure}
    \centering
    \includegraphics[width=0.48\textwidth]{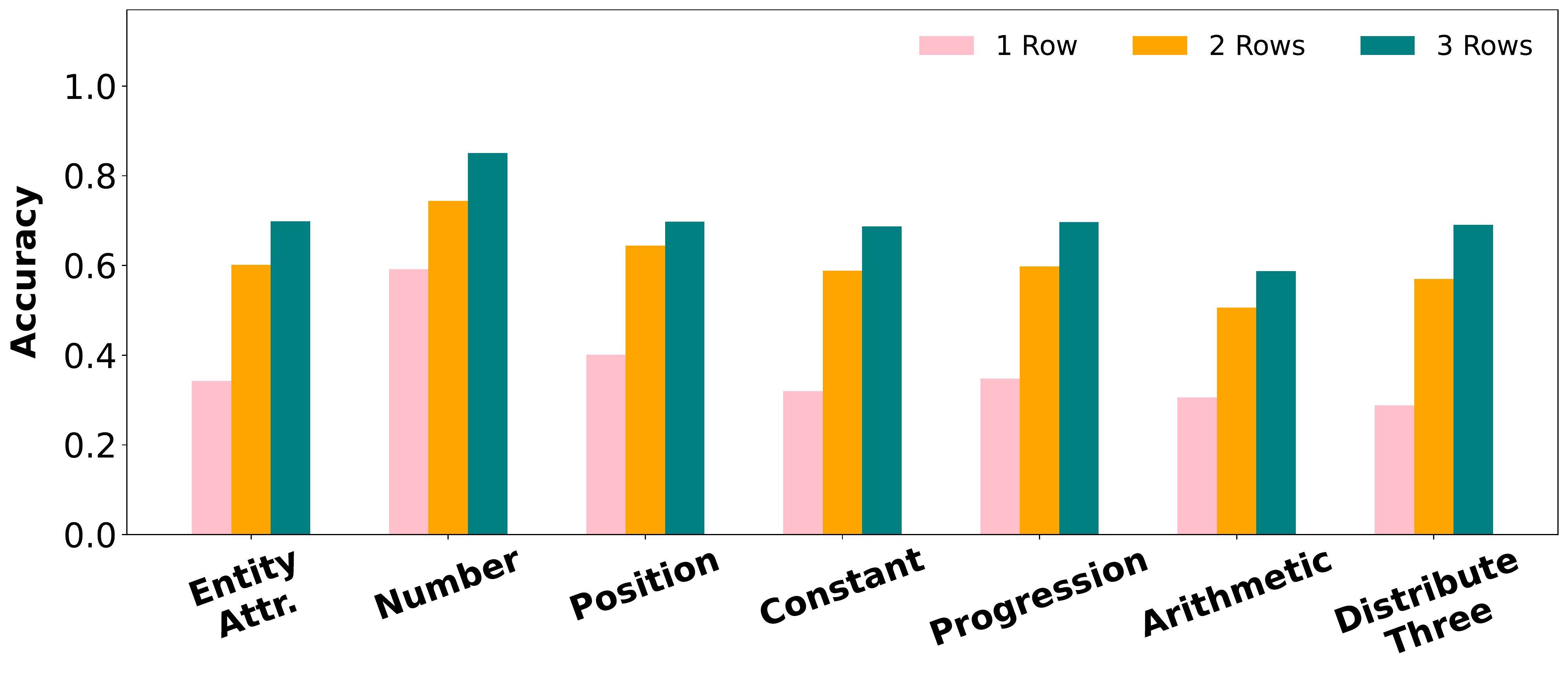}
    \vspace{-22pt}
    \caption{Comparison of accuracy on examples from all RAVEN sub-tasks as rows are introduced to the matrix, with \textbf{only entity attribute naming abstractions}.}
    \vspace{-5pt}
    \label{fig: in context learning analysis}
\end{figure}

\begin{figure}
    \centering
    \includegraphics[width=0.48\textwidth]{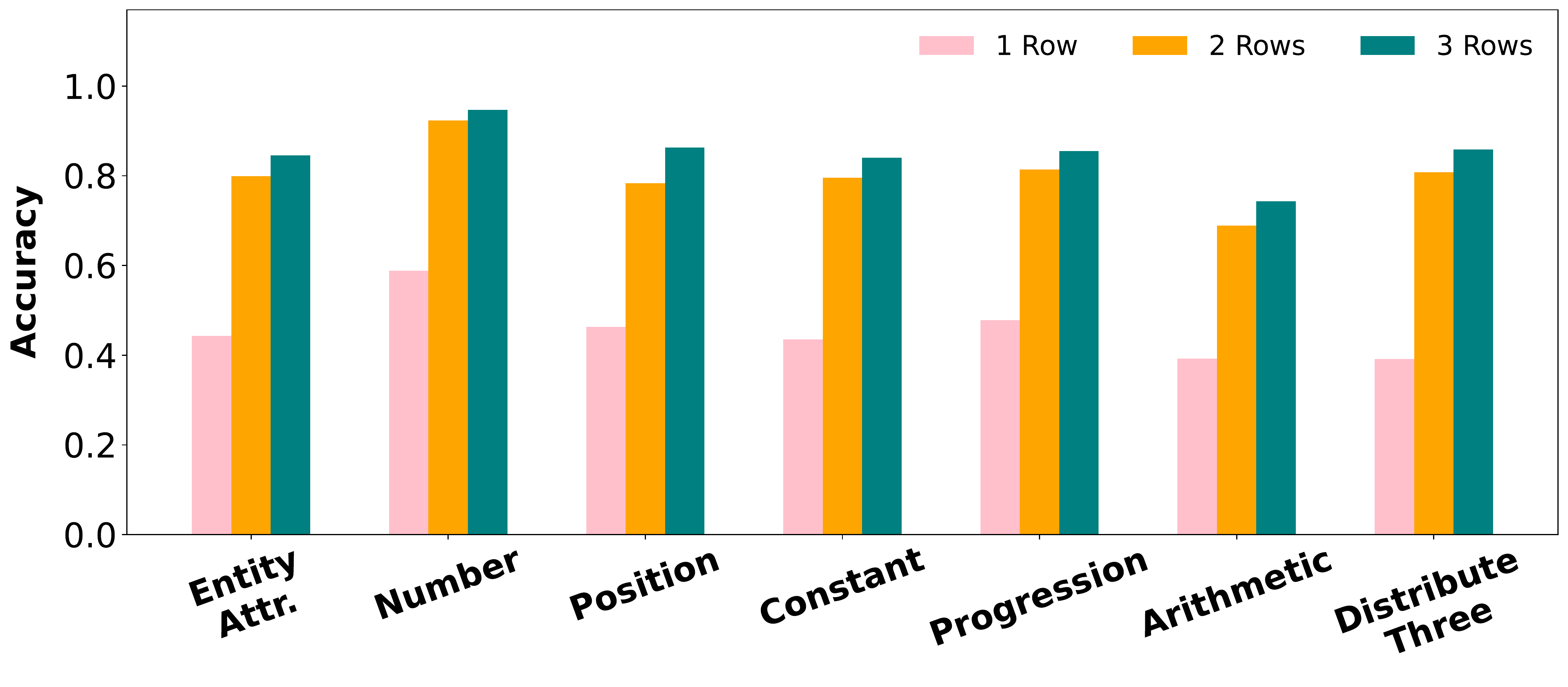}
    \vspace{-22pt}
    \caption{Comparison of accuracy on examples from all RAVEN sub-tasks as rows are introduced to the matrix, with \textbf{all decomposition abstractions}.}
    \vspace{-15pt}
    \label{fig: in context learning analysis 2}
\end{figure}

\paragraph{In-context learning of attributes and relations.}
{We may wonder whether specific relations or attributes are easier to understand than others with less context. For example, the \texttt{Progression} or \texttt{Constant} relations may be possible to recognize only from the first two items of the third row in an RPM, as we can easily observe patterns in attribute values here, e.g., that entity \texttt{size} is increasing or \texttt{color} remains constant. 
In Figures~\ref{fig: in context learning analysis} and \ref{fig: in context learning analysis 2}, we surprisingly observe only marginal differences here, except for the \texttt{number} attribute, which seems significantly better captured than other attributes in this no-context setting.
}

\section{Conclusion}

In this work, we explored the ability of large PLMs to perform zero-shot analogical reasoning in visual Raven's Progressive Matrices (RPM). Upon the simplest mapping to language, they can achieve striking results, while applying higher-level naming and decomposition abstractions over the task features further raises performance to the level of humans and supervised approaches in some cases. We find that while ordinal naming abstractions are a powerful way to enable analogical reasoning in larger PLMs, decomposition abstractions that break the task down into atomic parts conserve their working memory such that even smaller PLMs under 1B parameters can achieve competitive performance on this challenging problem. 

Our detailed analysis revealed insights about which features of the task PLMs best capture, their robustness to distracting features, and the role of in-context learning and prior knowledge in picking up this complex task.
Surprisingly, we find that even without two complete rows of prior context from the matrix, GPT-3 175B and smaller models can achieve above-chance performance on the task, raising questions about the objectivity and true role of prior knowledge in RPM tasks, which are assumed to require minimal prior knowledge.

These results also raise some questions about the role PLMs may play in future AI systems capable of analogy. While previously thought to be a difficult problem for AI systems, PLMs can solve the reasoning step of analogy easily given strong abstractions over visual perception. Many of these abstractions are intuitive and commonly researched in computer vision, including the detection of object types, sizes, colors, counts, and global arrangements.
As such, future work may dive deeper into the challenging problem of generalized perception across domains, where we must robustly tease apart the key features of tasks and experiences that may facilitate analogy-making, e.g., in recognizing the commonalities between a physical bridge and the bridge of a song \cite{mitchell2021abstraction}.
Recent efforts toward understanding how humans describe abstract visual features in language by mapping them to natural concepts\footnote{For example, when communicating about abstract shapes, we may make an analogy to refer to them as looking like more familiar natural concepts like flowers or dog bones.} are a promising direction toward this goal \cite{draw-me-a-flower-2022,ji-2022-tangrams}.

\section*{Acknowledgements}
This work was supported in part by DARPA PTG program HR00112220003. We would like to thank the anonymous reviewers for their valuable comments and suggestions.

\section*{Limitations}

\paragraph{Perception and reasoning in text-based RAVEN.}
In this work, one limitation is that we do not attempt to solve the perception problem of analogy-making in RPM, rather we apply perfect perception in solving the reasoning part, and assume the perception problem is simple. By doing so, we find that PLMs may be a strong solution to the reasoning problem here, which may better direct future efforts toward AI and analogy. Obviously, the perception problem for idealized domains is a lot different than more natural domains, and identifying key features across many domains that can facilitate a mapping is still a challenging unsolved problem. We hope that our work sparks more interest in this problem.

Meanwhile, one may argue that our decomposition abstractions are too strong, and actually contribute to the reasoning problem in RPM, as they make an independence assumption about which features of the task can be teased apart. Making such an assumption requires an understanding of the problem that cannot be inferred by only seeing one instance. However, we decomposed the task based on very intuitive and common attributes, e.g., shapes, colors, sizes, and counts of items. We believe that the strength of such an abstraction, which could be applied in many problems, should not be understated.
Nonetheless, we include decomposition-free forms of results as much as possible throughout the paper to help compare the contributions of decomposition versus naming abstractions, which is more clearly only providing perceptual information. In fact, we find that without any decomposition, PLMs still achieve very strong performance in many cases, and performance gains from decomposition are not always large.

\paragraph{Human performance.}
{Lastly, we note some limitations in the human performance measurements used as reference points. In \citet{zhang2019raven}, human performance on RAVEN was measured by giving subjects some task-specific training, then evaluating them on the original visual form of the task. This differs from our results in two ways. First, PLMs had no task-specific training for RAVEN, given that experiments were zero-shot and the text data we generate is new and thus impossible to appear directly in PLM pre-training. This may give humans an advantage. Second, the task is presented to PLMs in text form, not visually. While the essential information from the task is preserved by our conversion, it is possible that this conversion would affect the difficulty of the task for humans (making it easier or harder). As such, it becomes unclear how to contextualize our results with these past human results. Future work may carry out systematic human studies to compare the analogical reasoning capabilities of humans and PLMs in different settings.}

\section*{Ethical Considerations}
This work does not use any human subjects or human-generated data. 
Our work deals with abstract visual features that are described with numerical symbols, thus not strongly targeting any language. A possible ethical concern for this work is the amount of computational resources used in evaluating PLMs. To reduce unnecessary computation in our study, we chose to apply PLMs to only a subset of 500 testing examples from each sub-task of the RAVEN dataset, while the full testing set is four times as large.

\bibliography{custom}

\clearpage
\appendix

\section{Expanded Results}
\label{apx: expanded results}

In Table~\ref{tab:expanded results}, we present additional results with a wider range of OPT model sizes~\cite{zhang2022opt}. We observe similar mostly monotonic increases of accuracy with model size.

\section{Results and Analysis with I-RAVEN}\label{apx: iraven results}
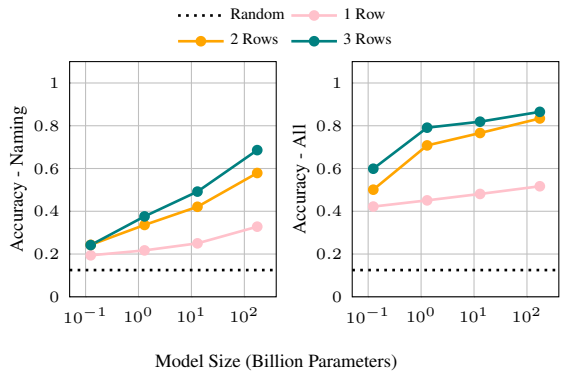
\begin{figure}[h]
\centering
\vspace{-7pt}
\begin{tikzpicture}
\begin{groupplot}[group style={group size=2 by 1, horizontal sep=1cm}, width=0.2\textwidth, legend style={draw=none, anchor=north, legend columns=2}, legend to name=commonlegend, legend cell align=left, xlabel={}, xmode=log, xtick={0.1,1,10,100}, ytick={0,0.2,0.4,0.6,0.8,1}, minor tick style={draw=none}, xtick pos=bottom, xtick align=outside, ytick pos=left, ytick align=outside, major tick length=0, xmin=0.05, xmax=400, ymin=0, ymax=1.1, compat=1.18, xmajorgrids=true, ymajorgrids=true]

\nextgroupplot[ylabel={\scriptsize Accuracy - Naming}, ylabel style={yshift=-5pt}, xlabel={\scriptsize Model Size (Billion Parameters)}, xlabel style={at={(1,-0.2)}}, width=4.3cm, height=4.7cm]

\addplot[mark=*,mark size=1.5pt,color=pink,line width=1pt] coordinates {(0.125, 0.194) (1.3, 0.217) (13, 0.25) (175, 0.328)};
\addlegendentry{\tiny\color{black}1 Row}

\addplot[mark=*,mark size=1.5pt,color=orange,line width=1pt] coordinates {(0.125, 0.243) (1.3, 0.336) (13, 0.421) (175, 0.579)};
\addlegendentry{\tiny\color{black}2 Rows}

\addplot[mark=*,mark size=1.5pt,color=teal,line width=1pt] coordinates {(0.125, 0.242) (1.3, 0.376) (13, 0.492) (175, 0.686)};
\addlegendentry{\tiny\color{black}3 Rows}

\addplot[color=black,dotted,line width=1pt] coordinates {(0.05, 0.125) (400, 0.125)};
\addlegendentry{\tiny Random}

\nextgroupplot[ylabel={\scriptsize Accuracy - All}, ylabel style={yshift=-5pt}, width=4.3cm, height=4.7cm]

\addplot[color=black,dotted,line width=1pt] coordinates {(0.05, 0.125) (400, 0.125)};
\addlegendentry{\tiny\color{black}Random}

\addplot[mark=*,mark size=1.5pt,color=pink,line width=1pt] coordinates {(0.125, 0.422) (1.3, 0.451) (13, 0.481) (175, 0.517)};
\addlegendentry{\tiny\color{black}1 Row}

\addplot[mark=*,mark size=1.5pt,color=orange,line width=1pt] coordinates {(0.125, 0.501) (1.3, 0.708) (13, 0.766) (175, 0.834)};
\addlegendentry{\tiny\color{black}2 Rows}

\addplot[mark=*,mark size=1.5pt,color=teal,line width=1pt] coordinates {(0.125, 0.599) (1.3, 0.791) (13, 0.819) (175, 0.865)};
\addlegendentry{\tiny\color{black}3 Rows}

\end{groupplot}
\node at ($(group c1r1.center)+(1.5,2cm)$) {\ref{commonlegend}};
\end{tikzpicture}
\vspace{-10pt}
\caption{Macro average accuracy over all Impartial-RAVEN sub-tasks as we introduce rows to the matrix during in-context learning, under naming abstractions only (left) and all naming and decomposition abstractions (right). In 1 Row, we include only the incomplete third row.}
\label{fig: prior graph iraven}
\end{figure}

As the generation strategy for the negative choices in RAVEN can introduce distributional bias that is problematic for supervised learning and leads to artificially high performance~\cite{hu2021stratified}, this could be a possible reason behind PLMs' strong performance on the task even without any complete rows of context. As such, in 
Table~\ref{tab:expanded results 2} and Figure~\ref{fig: prior graph iraven}, we include some supplementary analysis on the Impartial-RAVEN (I-RAVEN) dataset from \citeauthor{hu2021stratified}, which introduces more variation in negative choices. However, we observe similar performance trends in I-RAVEN. Performance mostly monotonically increases with model sizes and more abstraction. Further, PLMs achieve above-chance performance again without any rows of context provided, even with no decomposition abstractions. This provides further evidence that RPM, at least formulated in this way, is in part addressed by PLMs' prior knowledge, despite the assumptions of minimal background knowledge that the task makes.

\begin{table*}[!ht]
    \small
    \centering
    \setlength{\columnsep}{1pt}
    \begin{tabular}{c c | c c c c c c c | c}
        \toprule
        ~ & \textbf{Abstractions} & \texttt{\textbf{Center}} & \texttt{\textbf{2x2}} & \texttt{\textbf{3x3}} & \texttt{\textbf{L-R}} & \texttt{\textbf{U-D}} & \texttt{\textbf{O-IC}} & \texttt{\textbf{O-IG}} & \textbf{Avg.} \\ \midrule
        
        \multirow{3}{*}{125M} & Attr. Naming Only & 0.222 & 0.420 & 0.606 & 0.076 & 0.098 & 0.122 & 0.194 & 0.248 \\ 
        ~ & Comp. Decomp. & 0.222 & 0.420 & 0.606 & 0.136 & 0.154 & 0.162 & 0.222 & 0.275 \\ 
        ~ & Comp. + Attr. Decomp. & 0.456 & 0.620 & 0.724 & 0.378 & 0.408 & 0.374 & 0.520 & 0.497 \\ \midrule
        \multirow{3}{*}{350M} & Attr. Naming Only & 0.302 & 0.510 & 0.684 & 0.104 & 0.134 & 0.120 & 0.250 & 0.301 \\ 
        ~ & Comp. Decomp. & 0.302 & 0.510 & 0.684 & 0.186 & 0.232 & 0.254 & 0.344 & 0.359 \\ 
        ~ & Comp. + Attr. Decomp. & 0.436 & 0.588 & 0.788 & 0.280 & 0.346 & 0.290 & 0.408 & 0.448 \\ \midrule
        \multirow{3}{*}{1.3B} & Attr. Naming Only & 0.472 & 0.584 & 0.710 & 0.146 & 0.158 & 0.2 & 0.322 & 0.370 \\ 
        ~ & Comp. Decomp. & 0.472 & 0.584 & 0.710 & 0.410 & 0.426 & 0.434 & 0.494 & 0.504 \\ 
        ~ & Comp. + Attr. Decomp. & 0.720 & 0.714 & 0.794 & 0.672 & 0.680 & 0.744 & 0.744 & 0.724 \\ \midrule
        \multirow{3}{*}{2.7B} & Attr. Naming Only & 0.534 & 0.572 & 0.746 & 0.216 & 0.2 & 0.268 & 0.336 & 0.410 \\ 
        ~ & Comp. Decomp. & 0.534 & 0.572 & 0.746 & 0.420 & 0.468 & 0.484 & 0.532 & 0.537 \\ 
        ~ & Comp. + Attr. Decomp. & 0.706 & 0.738 & 0.826 & 0.658 & 0.664 & 0.704 & 0.784 & 0.726 \\ \midrule
        \multirow{3}{*}{6.7B} & Attr. Naming Only & 0.618 & 0.590 & 0.752 & 0.196 & 0.228 & 0.284 & 0.396 & 0.438 \\
        ~ & Comp. Decomp. & 0.618 & 0.590 & 0.752 & 0.492 & 0.528 & 0.548 & 0.584 & 0.587 \\ 
        ~ & Comp. + Attr. Decomp. & 0.704 & 0.750 & 0.826 & 0.682 & 0.690 & 0.748 & 0.834 & 0.748 \\ \midrule
        \multirow{3}{*}{13B} & Attr. Naming Only & 0.644 & 0.610 & 0.754 & 0.220 & 0.268 & 0.358 & 0.452 & 0.472 \\
        ~ & Comp. Decomp. & 0.644 & 0.610 & 0.754 & 0.566 & 0.602 & 0.586 & 0.576 & 0.620 \\ 
        ~ & Comp. + Attr. Decomp. & 0.746 & 0.794 & 0.830 & 0.710 & 0.702 & 0.770 & 0.840 & 0.770 \\ \midrule
        \multirow{3}{*}{30B} & Attr. Naming Only & 0.680 & 0.596 & 0.748 & 0.264 & 0.328 & 0.420 & 0.482 & 0.503 \\ 
        ~ & Comp. Decomp. & 0.680 & 0.596 & 0.748 & 0.582 & 0.618 & 0.664 & 0.638 & 0.647 \\ 
        ~ & Comp. + Attr. Decomp. & 0.762 & 0.818 & 0.828 & 0.738 & 0.714 & 0.786 & 0.860 & 0.787 \\ \midrule
        \multirow{3}{*}{175B} & Attr. Naming Only & 0.772 & 0.780 & 0.864 & 0.542 & 0.536 & 0.648 & 0.748 & 0.699 \\
        ~ & Comp. Decomp. & 0.772 & 0.780 & 0.864 & 0.738 & 0.732 & 0.780 & 0.840 & 0.787 \\ \
        ~ & Comp. + Attr. Decomp. & 0.800 & 0.878 & 0.932 & 0.776 & 0.780 & 0.828 & 0.926 & 0.846 \\ \midrule
    \end{tabular}
    \caption{Performance on RAVEN sub-tasks under our abstractions across a wider set of model sizes. 175B refers to \texttt{text-davinci-002} while the rest are corresponding OPT models.}
    \label{tab:expanded results}
\end{table*}

\begin{table*}[!ht]
    \small
    \centering
    \setlength{\columnsep}{1pt}
    \begin{tabular}{{c c | c c c c c c c | c}}
    \toprule
        ~ & \textbf{Abstractions} & \texttt{\textbf{Center}} & \texttt{\textbf{2x2}} & \texttt{\textbf{3x3}} & \texttt{\textbf{L-R}} & \texttt{\textbf{U-D}} & \texttt{\textbf{O-IC}} & \texttt{\textbf{O-IG}} & \textbf{Avg.} \\ \midrule
        \multirow{3}{*}{125M} & Attr. Naming Only & 0.376 & 0.172 & 0.208 & 0.246 & 0.230 & 0.262 & 0.202 & 0.242 \\
        ~ & Comp. Decomp. & 0.376 & 0.172 & 0.208 & 0.336 & 0.344 & 0.354 & 0.224 & 0.288 \\
        ~ & Comp. + Attr. Decomp. & 0.608 & 0.514 & 0.602 & 0.612 & 0.624 & 0.638 & 0.594 & 0.600 \\ \midrule
        \multirow{3}{*}{1.3B} & Attr. Naming Only & 0.594 & 0.290 & 0.310 & 0.348 & 0.370 & 0.388 & 0.334 & 0.376 \\
        ~ & Comp. Decomp. & 0.594 & 0.290 & 0.310 & 0.586 & 0.574 & 0.618 & 0.466 & 0.491 \\
        ~ & Comp. + Attr. Decomp. & 0.810 & 0.676 & 0.730 & 0.822 & 0.802 & 0.882 & 0.818 & 0.791 \\ \midrule
        \multirow{3}{*}{13B} & Attr. Naming Only & 0.756 & 0.384 & 0.382 & 0.456 & 0.498 & 0.538 & 0.432 & 0.492 \\
        ~ & Comp. Decomp. & 0.756 & 0.384 & 0.382 & 0.750 & 0.74 & 0.766 & 0.564 & 0.620 \\
        ~ & Comp. + Attr. Decomp. & 0.836 & 0.748 & 0.728 & 0.824 & 0.826 & 0.906 & 0.868 & 0.819 \\ \midrule
        \multirow{3}{*}{175B} & Attr. Naming Only & 0.808 & 0.564 & 0.566 & 0.656 & 0.676 & 0.818 & 0.714 & 0.686 \\
        ~ & Comp. Decomp. & 0.808 & 0.564 & 0.566 & 0.822 & 0.812 & 0.896 & 0.742 & 0.744 \\
        ~ & Comp. + Attr. Decomp. & 0.864 & 0.832 & 0.818 & 0.834 & 0.846 & 0.928 & 0.930 & 0.865 \\ \midrule
    \end{tabular}
    \caption{Performance on I-RAVEN sub-tasks under our abstractions across different model sizes. 175B refers to \texttt{text-davinci-002} while the rest are corresponding OPT models.}
    \label{tab:expanded results 2}
\end{table*}

\section{Example Prompts}\label{apx: example prompts}

\begin{figure*}
    \centering
    \includegraphics[width=0.93\textwidth]{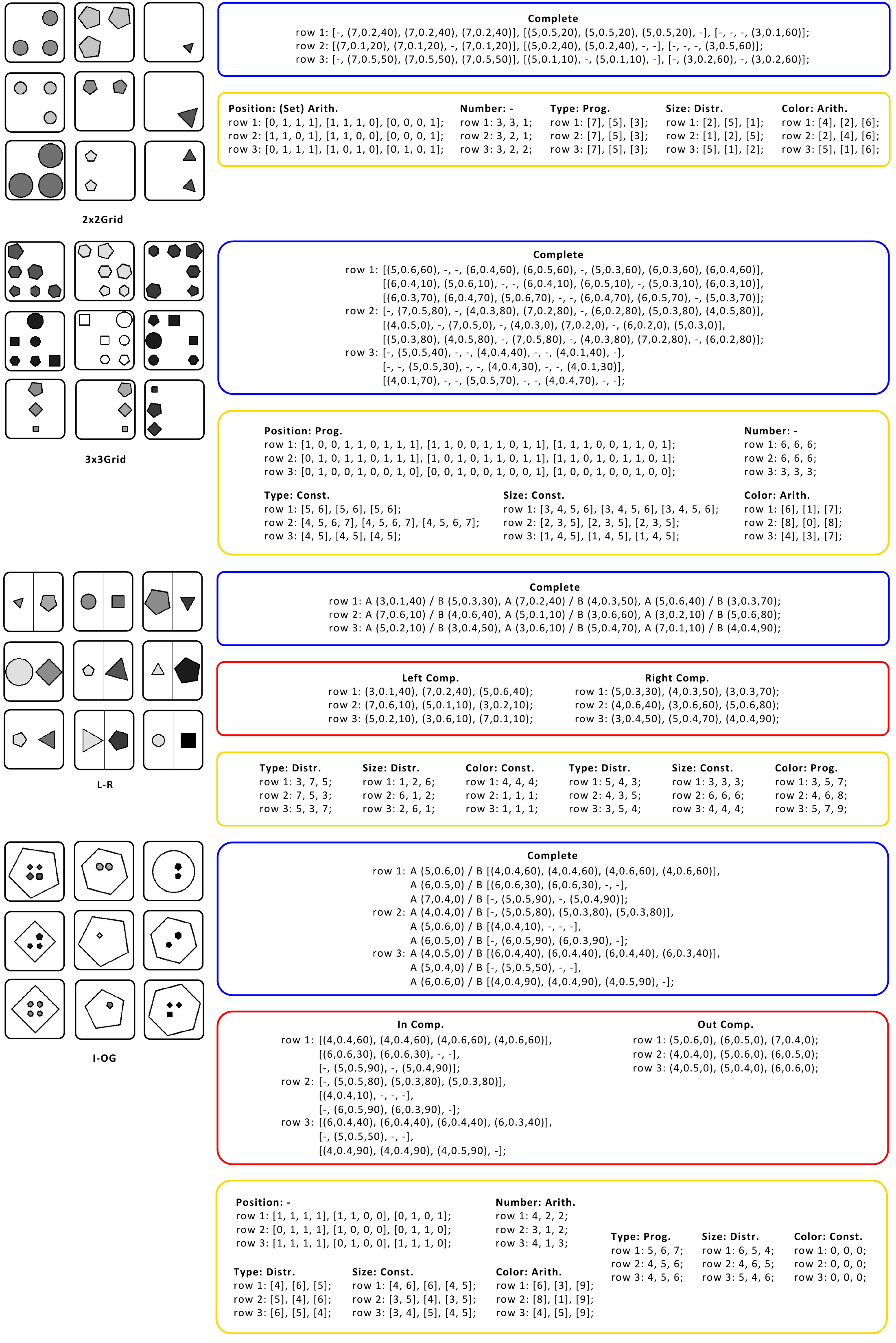}
    \caption{Example prompts for \texttt{2x2Grid}, \texttt{3x3Grid}, \texttt{L-R} and \texttt{I-OG} subtasks under different abstractions.}
    \label{fig: example prompts}
\end{figure*}

In Figure \ref{fig: example prompts}, we include example prompts for \texttt{2x2Grid}, \texttt{3x3Grid}, \texttt{L-R} and \texttt{I-OG} subtasks under different abstractions. Note that \texttt{U-D} and \texttt{I-OC} are isomorphic to \texttt{L-R}, and therefore share the same prompt format.

\end{document}